\documentclass[journal]{IEEEtran}

\usepackage{amsmath,amsfonts}
\usepackage{algorithmic}
\usepackage{array}
\usepackage{textcomp}
\usepackage{stfloats}
\usepackage{url}
\usepackage{verbatim}
\usepackage{graphicx}
\usepackage{balance}

\usepackage{booktabs}
\usepackage{threeparttable}

\usepackage{cite}

\usepackage[autolanguage]{numprint}

\usepackage{subcaption}

\usepackage{xcolor}

\usepackage[acronym]{glossaries}

\usepackage{hyperref}
\usepackage[capitalise]{cleveref}
\crefname{figure}{Fig.}{Figs.}
\Crefname{figure}{Fig.}{Figs.}
\crefname{table}{Table}{Tables}
\Crefname{table}{Table}{Tables}

\newacronym{dnn}{DNN}{deep neural network}
\newacronym{qat}{QAT}{quantisation-aware training}
\newacronym{mi}{MI}{mutual information}
\newacronym{ip}{IP}{information plane}
\newacronym{ib}{IB}{information bottleneck}
\newacronym{ste}{STE}{straight-through estimator}
\newacronym{bnn}{BNN}{binary neural network}
\newacronym{fc}{FC}{fully connected}

\DeclareMathOperator{\sgn}{sign}

\newacronym{sgd}{SGD}{stochastic gradient descent}
\newacronym{erm}{ERM}{empirical error minimisation}
\newacronym{relu}{ReLU}{rectified linear unit}
\newacronym{dpi}{DPI}{data processing inequality}
\newacronym{qnn}{QNN}{quantised neural network}
\newacronym{szt}{SZT}{Shwartz-Ziv-Tishby}
\newacronym{pmf}{PMF}{probability mass function}
\newacronym{iid}{iid}{independent and identically distributed}
\newacronym{rv}{RV}{random variable}
\newacronym{bn}{BN}{batch normalisation}
\newacronym{cv}{CV}{convolution}
\newacronym{mp}{MP}{max pooling}

\begin{document}

\title{Information Plane Analysis of Binary Neural Networks}

\author{
    Maximilian Nothnagel and
    Bernhard C. Geiger, \IEEEmembership{Senior Member, IEEE}
    \thanks{M. Nothnagel (\texttt{nothnagel.maximilian@proton.me}) is with Graz University of Technology, Graz, Austria and was with Know Center Research GmbH, Graz, Austria.}
    \thanks{B. C. Geiger (\texttt{geiger@tugraz.at}) is with the Signal Processing and Speech Communication Laboratory, Graz University of Technology, Graz, Austria, with Know Center Research GmbH, Graz, Austria, and with the Graz Center for Machine Learning (GraML), Graz, Austria. }
}

\markboth{IEEE Transactions on Pattern Analysis and Machine Intelligence,~Vol.~AA, No.~BB, MMMM~YYYY}%
{Information Plane Analysis of Binary Neural Networks}

\maketitle

\begin{abstract}
    Information plane (IP) analysis has been suggested to study the training dynamics of deep neural networks through mutual information (MI) between inputs, representations, and targets. However, its statistical validity is often compromised by the difficulty of estimating MI from samples of high-dimensional, deterministic representations. 
    In this work, we perform IP analyses on binary neural networks (BNNs) where activations are discrete and MI is finite. We characterise the finite-sample behaviour of the plug-in entropy estimator and identify regimes for sample size $N$ and representation dimensionality $D$ under which MI estimates are reliable. Outside these regimes, we show that empirical MI estimates saturate to $\log_2 N$, rendering IP trajectories uninformative.
    Restricting attention to the reliable regime, we train 375 BNNs to investigate the existence of late-stage compression phases and the relationship between compressed representations and generalisation performance. Our results show that while late-stage compression is frequently observed, compressed latent representations do not consistently correlate with improved generalization performance. Instead, the relationship between compression and generalisation is highly dependent on task, architecture, and regularisation.
\end{abstract}


\section{Introduction}

\IEEEPARstart{B}{ased} on the \gls{ib} theory, the authors of \cite{shwartz-ziv_opening_2017} introduce the concept of \gls{ip} analysis of \glspl{dnn}. Given the latent representations $T_\ell$ of a neuron layer $\ell$, the \gls{mi} $I(X;T_\ell)$ between input and representation  and the \gls{mi} $I(T_\ell;Y)$ between target and representation are plotted for each epoch on respective axes. The authors claim that \glspl{ip} can \textit{visually} explain the inner training dynamics of \glspl{dnn}, by analysing \textit{compression} and \textit{fitting} phases---i.e., the decrease of $I(X;T_\ell)$ and increase of $I(T_\ell;Y)$, respectively~\cite{shwartz-ziv_opening_2017,geiger_information_2022}.

The concept of \gls{ip} analysis was met with a broad number of subsequent analyses by several authors; however, the evidence regarding the importance or even existence of the compression phase has been inconclusive and topic of discussion~\cite{geiger_information_2022,lorenzen_information_2022}. While initially the existence of a compression phase has been linked to the usage of certain activation functions (e.g.,~\cite{saxe_information_2019}), more recent work has argued that the qualitative picture in the IP is heavily influenced by the properties of the method of estimating \gls{mi}. Specifically, for deterministic \glspl{dnn}, where \gls{mi} is infinite for a continuous-valued input~\cite[Th.~1]{Amjad_HowNotTo}, the \gls{ip} is now claimed to show a geometric, rather than information-theoretic picture due to the inherent properties of many estimators~\cite{geiger_information_2022}, including the popular binning estimator~\cite{goldfeld_estimating_2018}. Furthermore, even if true \gls{mi} was bounded, the performance of many estimators depends on the dimensionality of the latent representation $T_\ell$ and the number $N$ of samples available for estimation. Taken together, these considerations render many of previous \gls{ip} analyses unreliable, even for questions as simple as regarding the existence of a compression phase.

In this work, we present \gls{ip} analyses free from these shortcomings. First, we focus on \glspl{bnn}~\cite{courbariaux_binaryconnect_2015,hubara_quantized_2018}, in which the neuron activations are inherently discrete values and where, hence, \gls{mi} terms are finite. For these discrete activations, the plug-in estimate~\cite{antos_convergence_2001} is an appropriate choice, but still affected by the dimensionality of $T_\ell$ and the availability of samples. Presenting previous theoretical results together with experiments on synthetic data with known entropy, we show in~\cref{sec:method:entropy} that entropy estimates outside a suggested data regime strongly diverge from the truth, as estimates start saturating for increased dimensionality.

Having thus established sufficient conditions for a reliable \gls{ip} analysis, our main contribution is an in-depth investigation of the interplay between information-theoretic compression and generalisation performance on well-chosen toy examples involving \glspl{bnn}. We are interested in the following questions: 
\begin{enumerate}
    \item Do \glspl{bnn} exhibit a compression phase in the \gls{ip} in the sense that the \gls{mi} $I(X;T_\ell)$ decreases during a late period in training (\cref{sec:latestage})?
    \item Do compressed representations, i.e., latent $T_\ell$ with small $I(X;T_\ell)$, correlate with improved generalisation performance (\cref{sec:compression})?
\end{enumerate} 
While the first question is connected to the ongoing debate about whether training \glspl{dnn} proceeds in two phases, the second question contributes to the practically relevant hypothesis that information-theoretic compression is a possible approach to prevent overfitting.

To answer above questions, we conduct 125 different experiments and three runs per experiment, totalling to 375 trained \glspl{bnn}, distributed among four datasets. We estimate the \gls{mi} quantities of the binary activations of model layers for which estimation can be done reliably. We conclude that compression is not necessarily indicating better generalisation. However, the movement and behaviour in the \gls{ip} can provide insight into the performance of the model. We discuss these and other implications of our work as well as its limitations in \cref{sec:discussion}.

\section{Related Work}

Several authors with varying approaches and focus have conducted \gls{ip} analyses on binarised or quantised neural networks. The authors of \cite{nguyen-tang_markov_2019} focused on stochastic \glspl{bnn} with four hidden layers, estimating the \gls{mi} using Monte Carlo samples. Experiments on the synthetic SZT dataset provided in \cite{shwartz-ziv_opening_2017} and MNIST reveal no compression phase. In \cite{raj_understanding_2020}, fully connected binarised networks were trained on SZT and MNIST, applying the plug-in estimate and binning the real valued input $X$ and softmax output $\hat{Y}$ into 30 bins. The authors observed compression for the binary activations for MNIST, but no `explicit' compression phase on SZT. \Gls{qat}~\cite{jacob_quantization_2018} was applied in \cite{lorenzen_information_2022} on different models. On SZT, ReLU and $\tanh$ as activation function with 4-, 8-, and 32-bit quantisation is applied, exhibiting compression in hidden layers only for $\tanh$. On MNIST, ReLU activation with 8-bit quantisation showed small compression.

Thus, the question regarding information-theoretic compression in \glspl{bnn} remains inconclusive, similar to general \glspl{dnn}. In addition, previous studies focus on the SZT and MNIST dataset, with a lack of experiments on more complex datasets and more varying architectures. Furthermore, previous work often analysed layer widths that lie outside a suggest samples-to-dimensionality regime (\cref{sec:method:entropy}), resulting in unreliable entropy and \gls{mi} estimation on either axis of the \gls{ip} plot.

\section{Methodology}

\subsection{Binary Neural Networks}

In this work, we apply the sign function \begin{equation}
    \label{eq:methods:bnn:sign}
    \sgn\,x = \begin{cases}
        1 & \text{if } x > 0, \\
        0 & \text{else}
    \end{cases}
\end{equation} to the activations of hidden \gls{fc} layers with batch normalisation in between. Input data and output of non-\gls{fc} layers are kept as full-precision real values. To circumvent the non-differentiability of the sign function, a \gls{ste}~\cite{hinton_neural_2012,bengio_estimating_2013} is applied during the backward pass. Let $t = \sgn z$ be the quantisation of the previous (pre-)activation $z$ and assume that the gradient $\frac{\partial C}{\partial t}$ has been obtained. Then, the saturation-aware \gls{ste}~\cite{hubara_quantized_2018} is defined as: \begin{equation}
    \label{eq:methods:bnn:ste}
    \frac{\partial C}{\partial z} = \frac{\partial C}{\partial t} \cdot \mathbf{1}_{|z| \leq 1},
    \quad \text{with } \mathbf{1}_{|z|\leq 1} = \begin{cases}
        1 & \text{if } -1 \leq z \leq 1, \\
        0 & \text{else}
    \end{cases}
\end{equation} The indicator function $\mathbf{1}_{|z| \leq 1}$ serves as a (hard) threshold regarding the magnitude of the non-quantised activation, cancelling the gradient if $|z|$ is too large~\cite{hubara_quantized_2018}.

\subsection{Entropy Estimation}\label{sec:method:entropy}

We utilise the discrete binary activations of our models and apply the empirical or plug-in estimate~\cite{antos_convergence_2001} for entropy. Let $X$ be a discrete \gls{rv} with an (unknown) probability mass function $p_X$ on an alphabet $\mathcal{X}$ of cardinality $k$. Given a set of $N$ \gls{iid} observations $\mathcal{D} = \left\{x_i\right\}_{i=1}^N$, where each $x_i$ is a realisation of $X$, the empirical distribution is computed as 
\begin{subequations}\label{eq:methods:ee:plugin}
\begin{equation}
    \label{eq:methods:ee:empirical-dist}
    \hat{p}_X(x) = \frac{n_x}{N},
\end{equation} where $n_x$ is the number of occurrences of $x$ in $\mathcal{D}$. `Plugging in' $\hat{p}_X$ into the definition of the entropy $H(X)$ of $X$, and defining $0\log 0=0$ by continuous extension, yields the plug-in estimate of entropy: \begin{equation}
    \label{eq:methods:ee:entropy}
    \hat{H}(X) = -\sum_{x \in \mathcal{X}} \hat{p}_X(x) \log_2 \hat{p}_X(x)
\end{equation}
\end{subequations}

The estimate is shown to be consistent for $N \to \infty$~\cite{antos_convergence_2001}. However, in the more practical `large $N$' regime, assuming $N$ to be a function of $k$, the estimate is suboptimal if the alphabet size is increased but the number of samples is not increased accordingly. More specifically, the estimate is suboptimal in the samples-to-dimensionality regime of $N=o\left(\frac{k^2}{\log_2^2 k}\right)$~\cite{wu_minimax_2016}.

Considering the datasets used in previous and our work, the ${N\approx800}$ samples in the validation set of the synthetic SZT data would suffice for alphabet sizes of up to $k\approx2^7$; for $N = \numprint{10000}$, that is MNIST and similar, the alphabet size can be as large as $k\approx2^{10}$. Assuming binary activations thus means that datasets of this size can reliable estimate entropy only in layers with up to seven (SZT) or ten (MNIST or similar) neurons.

\begin{figure*}[!t]
    \centering
    \includegraphics[width=\textwidth]{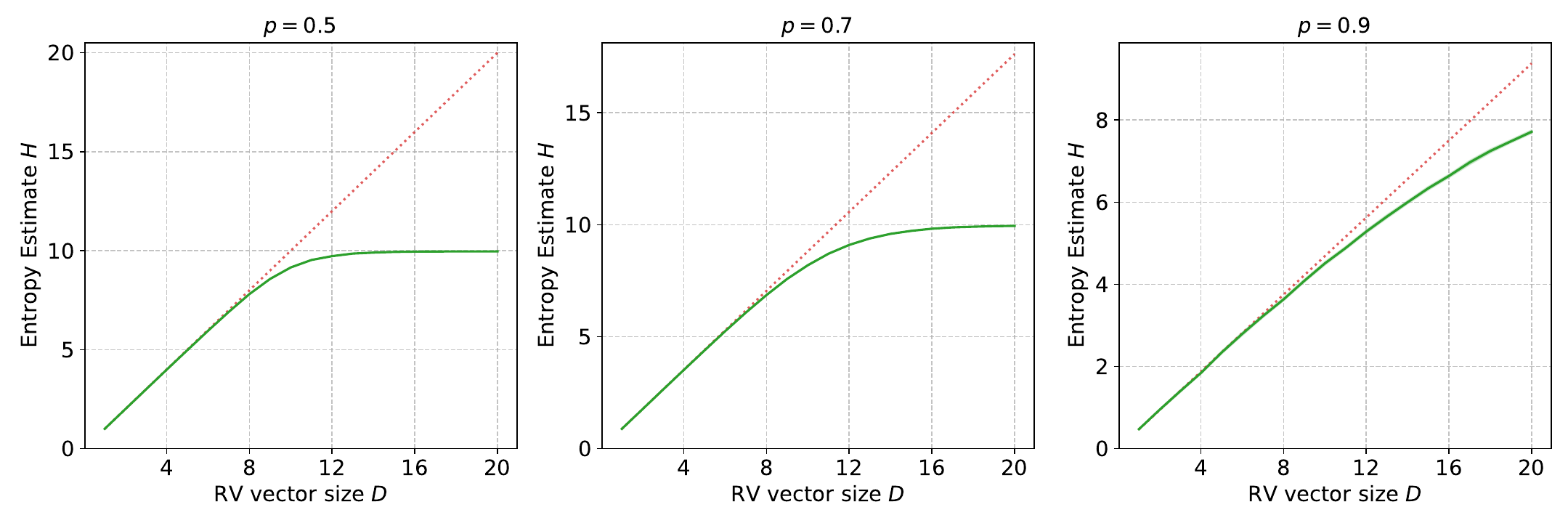}
    \caption{Evaluation of the plug-in estimate $\hat{H}$ with $N=\numprint{1000}$ samples of $D$-dimensional Bernoulli vectors of varying success probability $p$. Data is averaged over 20 experiments. The dotted line represents the true entropy $H$. It can be seen that while the true entropy increases linearly with dimensionality $D$, the estimate $\hat{H}$ starts diverging at $D \approx 8$ and eventually saturates at $\log_2 N$.}
    \label{fg:methods:ee:plug-in:suboptimality}
\end{figure*}

To illustrate this, \cref{fg:methods:ee:plug-in:suboptimality} shows an exemplary evaluation of the plug-in estimate on synthetic data. We generate 20 experiments with $N=\numprint{1000}$ samples of $D$-dimensional Bernoulli vectors $X=(X_1,\dots,X_D)$, with $D\in \{1,\dots,20\}$, where each vector dimension is drawn independently from a Bernoulli distribution with success probability $p \in \{0.5, 0.7, 0.9\}$. The true entropy per experiment is computed as: \begin{align}
    \begin{split}
        H(X) &= \sum_{d=1}^D H(X_d) = D \cdot h_2(p) \\
        &= -D \cdot \left(p \log_2(p) + q \log_2(q)\right), \quad q = 1-p
    \end{split}
\end{align} It is trivial to see that $H$ grows linearly with $D$. In addition, $\hat{H}$ starts underestimating at $D\geq 7$ for all $p$, adhering to the required regime proposition mentioned before. Beyond this limit, the estimate starts saturating to $\log_2 N$, as the alphabet size $k=2^D$ grows so large that each of the $N$ samples assumes a different value of the alphabet.

\subsection{Information Plane Analyses and Estimation of Mutual Information}

Treating $X$ and $Y$ as \glspl{rv}, the latent representation $T_\ell$ obtained by the binary activations of the $\ell$-th hidden layer is an \gls{rv} with an alphabet  $\mathcal{T}_\ell$ of cardinality $k_\ell = 2^{d_\ell}$, where $d_\ell$ is the number of neurons in this layer. For \gls{ip} analyses, estimates of the \gls{mi} terms $I(X;T_\ell)$ and $I(T_\ell;Y)$ are computed for each (or a relevant subset) of the training epochs. These estimates are computed from a validation set $\mathcal{D}=\{(x_i,y_i)\}_{i=1}^N$ of size $N$. More specifically, for a given epoch and a given layer index $\ell$, $\mathcal{D}_\ell=\{(t_{i},y_i)\}_{i=1}^N$ is the resulting set of realisations of the joint \gls{rv} $(T_\ell,Y)$, and $\mathcal{D}_{\ell,t} = \left\{t_i\right\}_{i=1}^N$, where $t_i \in \{0, 1\}^{d_\ell}$.

Since in the \glspl{bnn} in our study the latent representations $T_\ell$ are deterministic functions of $X$, we obtain~\cite{saxe_information_2019}
\begin{equation}
    \begin{split}
        I(X;T_\ell) = H(T_\ell) - H(T_\ell \mid X) = H(T_\ell).
    \end{split}
\end{equation}
Hence the estimate $\hat{I}(X;T_\ell)$ of $I(X;T_\ell)$ is obtained by setting
\begin{equation}
    \hat{I}(X;T_\ell) = \hat{H}(T_\ell)
\end{equation}
and by applying the plug-in estimator~\eqref{eq:methods:ee:plugin} to the dataset $\mathcal{D}_{\ell,t}$.

For the term $I(T_\ell;Y)$ we recognise that 
\begin{align}
        I(T_\ell;Y) 
        &= \sum_{t \in \mathcal{T}_\ell, y \in \mathcal{Y}} p_{T_\ell,Y}(t, y) \log_2 \frac{p_{T_\ell,Y}(t, y)}{p_{T_\ell}(t) \cdot p_Y(y)}
\end{align} 
and hence obtain \begin{equation}
    \hat{I}(T_\ell;Y) = \sum_{t \in \mathcal{T}_\ell, y \in \mathcal{Y}} \hat{p}_{T_\ell, Y}(t, y) \log_2 \frac{\hat{p}_{T_\ell, Y}(t, y)}{\hat{p}_{T_\ell}(t) \cdot \hat{p}_Y(y)},
\end{equation} 
where the empirical distributions $\hat{p}_{T_\ell}$, $\hat{p}_Y$ and $\hat{p}_{T_\ell, Y}(t, y)$ are estimated from $\mathcal{D}_\ell$ using~\eqref{eq:methods:ee:empirical-dist}. For example, the empirical joint distribution $\hat{p}_{T_\ell, Y}$ is given as: 
\begin{equation}
    \hat{p}_{T_\ell, Y}(t, y) = \frac{1}{N} \sum_{i = 1}^N \mathbf{1}_{(t_i, y_i) = (t, y)}
\end{equation}

\section{Experimental Setups}\label{sec:experiments}

We conduct \gls{ip} analyses of \glspl{bnn} on four datasets. For each dataset, different neural networks---varying in architecture and regularisation in the form of weight decay---are trained over \numprint{3000} epochs. Each experiment is performed on three consecutive runs.\footnote{The behaviour in the \glspl{ip} is quite similar over all runs. Thus, we only show the first run of experiments for all \gls{ip} plots.} In total, we analyse 375 trained models, of which select results are presented.\footnote{The code is available online on \url{https://github.com/InformationPlanesDecompositions/entropy-estimation}.} The \glspl{mi} of the latent representations of our models are estimated on the validation subset of the respective dataset. The output layer is not considered for \gls{mi} estimation and \gls{ip} analysis.

\subsection{Datasets}

We employ the synthetic SZT dataset introduced in \cite{shwartz-ziv_opening_2017}, as well as MNIST, FashionMNIST, and CIFAR-10. The SZT dataset represents a binary classification problem from a 12-dimensional binary input and contains \numprint{4096} unique samples. We follow previous work~\cite{raj_understanding_2020, lorenzen_information_2022} and use \numprint[\%]{20} of the samples as validation split $\mathcal{D}$.

MNIST and FashionMNIST contain \numprint{60000} training and \numprint{10000} test samples, each being a $28 \times \numprint[px]{28}$ greyscale image representing a handwritten digit or clothing article, respectively, associated with one label from ten classes.

CIFAR-10 consists of \numprint{60000} samples of $32 \times \numprint[px]{32}$ colour images assigned to one of ten classes representing objects. The dataset is split into \numprint{50000} training and \numprint{10000} test samples, the latter containing exactly \numprint{1000} samples per class.

For MNIST, FashionMNIST, and CIFAR-10, we use the designated test split as the dataset $\mathcal{D}$ from which information-theoretic quantities are estimated.

\subsection{Model Architecture and Regularisation}

In each experiment, all \gls{fc} layers of the neural networks are binary, apart from the output layer. The weights are kept at full precision. As activation function, the sign function in~\eqref{eq:methods:bnn:sign} is used, while applying the saturation-aware \gls{ste} in~\eqref{eq:methods:bnn:ste} as backward approximation. For the experiments without regularisation, the Adam optimiser is used with a learning rate of \numprint{1e-4} for SZT and \numprint{1e-5} otherwise. The batch size is set to 64 for SZT and 256 otherwise.

The models used vary in the number of hidden layers and neurons. \Cref{tb:experiments:setup:architectures} gives an overview of the applied model structures and on which datasets they are used. Between each hidden layer, batch normalisation is applied. For the hourglass and bottleneck variants, the variable neuron width is set to $A \in \{2, 4, 6, 8, 10\}$. LeNet5 models use the conventional form of $\text{CV}(6)-\text{MP}-\text{CV}(16)-\text{MP}$ for the convolutional part. Each \gls{cv} uses a kernel size of 5, batch normalisation, and \gls{relu} activation, followed by a $2 \times 2$ \gls{mp} with stride 2. Except for CIFAR-10, the first \gls{cv} adds a padding of 2 to its input. The variant LeNet5 model varies the neuron widths of the \gls{fc} layers with $A \in \{50, 70, 120\}$ and $B \in \{20, 50\}$.

\begin{table*}[!t]
    \centering
    \begin{threeparttable}[b]
        \caption{Overview of the model architectures used in our experiments and on which datasets they are applied. $A$ and $B$ refer to architecture variations (see text).}
        \label{tb:experiments:setup:architectures}
        \begin{tabular}{l|c|c}
            \toprule
            Name & Architecture & Datasets\\
            \midrule
            SZT\tnote{*} & ${10-8-6-4}$ & SZT, MNIST \\
            Raj-like\tnote{\textdagger} & ${1024-20-20-20-10}$ & MNIST, FashionMNIST \\
            Hourglass & ${1024-20-10-A-10-20-10}$ & MNIST, FashionMNIST \\
            Bottleneck & ${1024-20-10-A-10}$ & MNIST, FashionMNIST \\
            Small \gls{bnn} & ${50-10-10}$ & MNIST, FashionMNIST \\
            LeNet5\tnote{\textdaggerdbl} & ${\text{LeNet5}-120-84-10}$ & MNIST, FashionMNIST, CIFAR-10 \\
            variant LeNet5 & ${\text{LeNet5}-A-B-10}$ & CIFAR-10 \\
            small LeNet5 & ${\text{LeNet5}-50-50-10}$ & CIFAR-10 \\
            \bottomrule
        \end{tabular}
        \begin{tablenotes}
            \item[*] Used in \cite{raj_understanding_2020} in accordance with \cite{shwartz-ziv_opening_2017}.
            \item[\textdagger] Used in \cite{raj_understanding_2020} without the last layer with width $10$.
            \item[\textdaggerdbl] The convolutional activations are not binary.
        \end{tablenotes}
    \end{threeparttable}
\end{table*}

As an experimental variant, regularisation is added in the form of the weight decay coefficient ${\lambda\in\{0.1,0.2,0.5,0.7,1,1.1,1.2,1.5,1.7,2\}}$, resulting in a total number of ten applied coefficients. For these variants, the setups remain identical apart from using the AdamW optimiser. In the \gls{ip} plots, the notation $\lambda = 0$ means there is \textit{no} weight decay applied and the Adam optimiser is used.

\subsection{Metrics}\label{sec:experiments:metrics}

As performance metric, we compute the classification accuracy on the validation set as percentage. In order to quantify compression, we define the compression factor \begin{equation}
    \varrho = \frac{\hat I(X;T_\ell)_{\max} - \bar{I}(X;T_\ell)_{50}}{\hat I(X;T_\ell)_{\max}},
    \label{eq:experiments:metrics:compression-factor}
\end{equation} where $\hat I(X;T_\ell)_{\max}$ is the maximum value over all epochs and $\bar{I}(X;T_\ell)_\text{50}$ is the mean of $\hat I(X;T_\ell)$ over the last 50 epochs. This factor dismisses any intermediate, temporary compression and focuses on the end-of-training behaviour of the models. Lastly, in order to quantify the correlation between the end-of-training $\hat I(X;T_\ell)$ and performance, we compute Spearman's rank correlation coefficient $r_s$ over the mean values of \gls{mi} and accuracy over the last 50 epochs.

\section{Late-Stage Compression in BNNs}\label{sec:latestage}

Late-stage compression, as experimentally observed by~\cite{shwartz-ziv_opening_2017} and others, is characterised by the \gls{mi} $I(X;T_\ell)$ or its estimate $\hat I(X;T_\ell)$ reducing during a second phase\footnote{We are aware that this second phase of training, during which $I(X;T_\ell)$ is purported to reduce, can assume a large portion of the total number of epochs. We abuse terminology and use the term ``late-stage'' to refer to this phase.} of training. The compression factor $\varrho$ is positive if such late-stage compression occurs, thus acts as a proxy for this phenomenon.

\begin{figure*}[!t]
    \centering
    \includegraphics[width=0.8\linewidth]{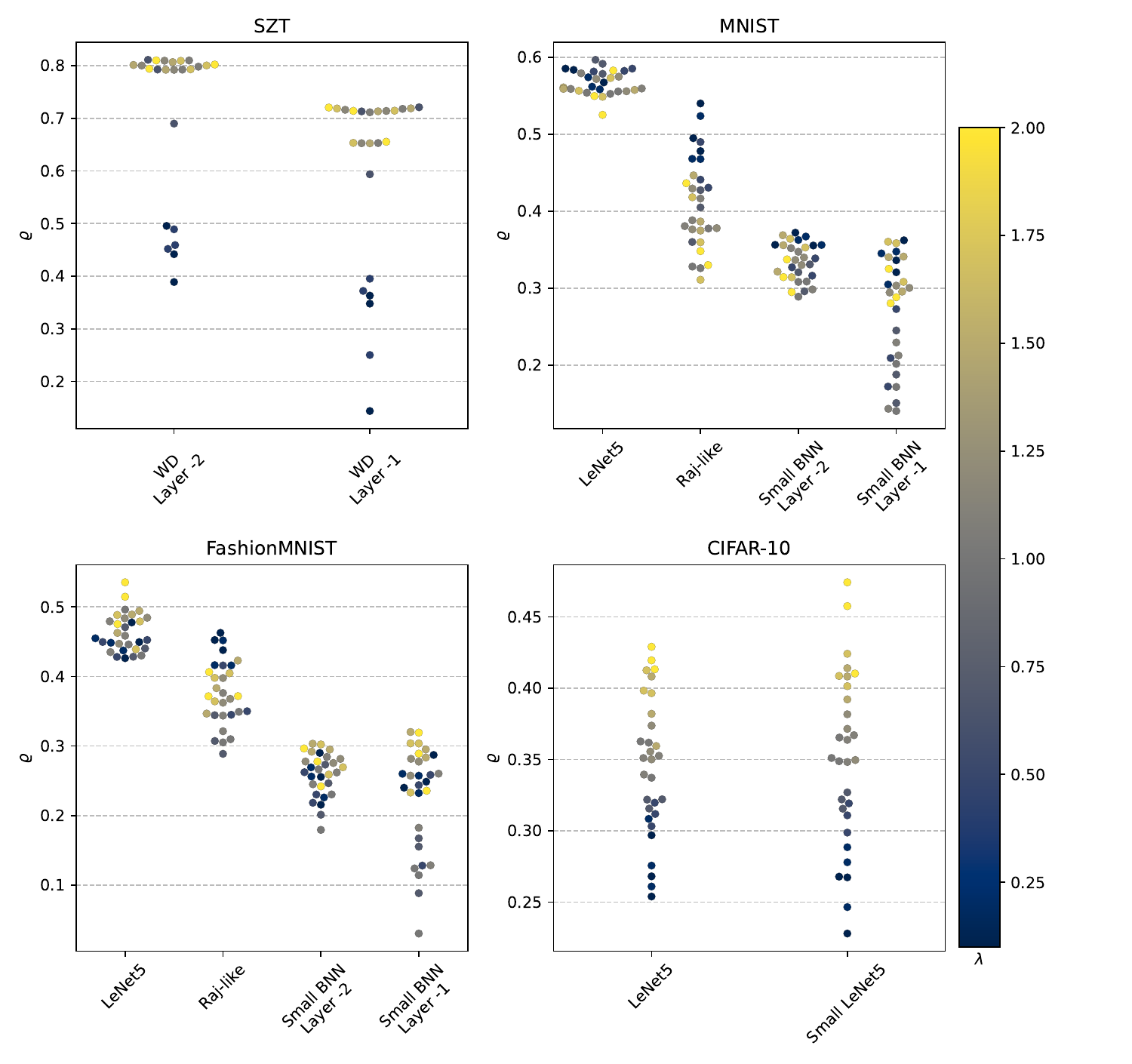}
    \caption{Compression factor $\varrho$ computed per experiment run and layer, grouped by dataset and experiment group. Each dot is coloured according to the weight decay coefficient $\lambda$ used in its experiment. Layer indices are given as negative offset starting from the output layer, i.e. layer -1 represents the last \textit{hidden} layer.}
    \label{fg:experiments:compression:compression-factor}
\end{figure*}

The compression factors for the weight decay experiments, obtained using~\eqref{eq:experiments:metrics:compression-factor}, are shown in \cref{fg:experiments:compression:compression-factor}. The factors are computed per run and layer and arranged per experiment and, if necessary, per layer. For all datasets and experiments, we observe compression to a certain degree, with a large majority of the  experiments exhibiting a compression factor of $\varrho \geq \numprint[\%]{25}$. Furthermore, regularisation in the form of weight decay can drive models to a stronger compression. This effect is not as clear for MNIST and FashionMNIST as it is for SZT and CIFAR-10. Later layers appear to compress less than earlier layers (as shown for the Small \gls{bnn} on MNIST and FashionMNIST), and compression in the last hidden layer seems to be more pronounced if the preceding part of the network has larger expressive power (LeNet5 vs. Small \gls{bnn}).

\begin{figure*}[!t]
    \centering
    \begin{subfigure}[t]{0.3\textwidth}
        \centering
        \includegraphics[height=2in]{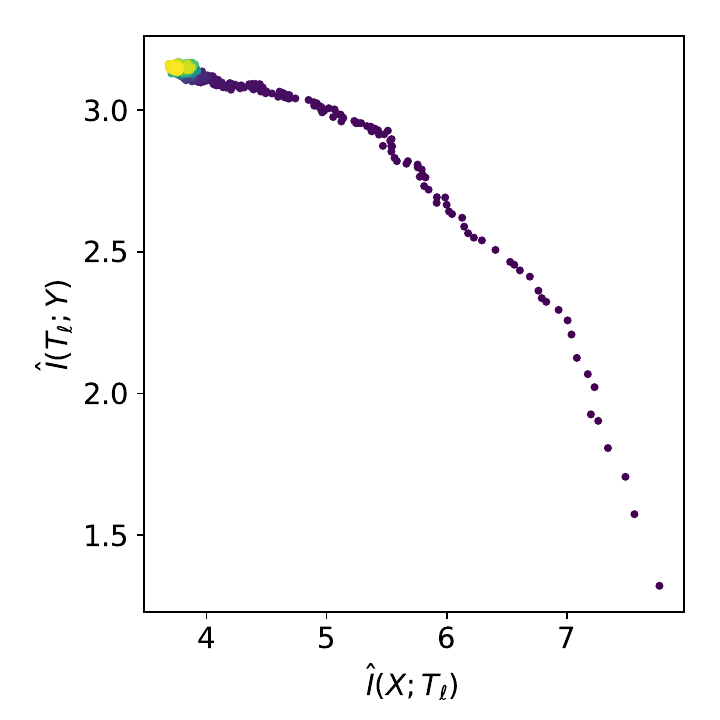}
        \caption{MNIST, Raj-like, $\lambda = 0$}
        \label{fg:experiments:compression:mnist}
    \end{subfigure}
    \begin{subfigure}[t]{0.3\textwidth}
        \centering
        \includegraphics[height=2in]{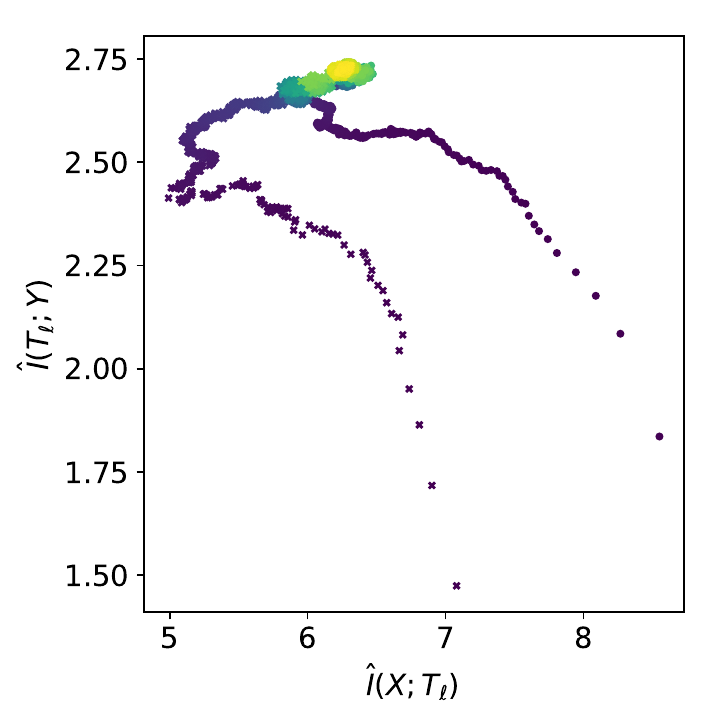}
        \caption{FashionMNIST, small \gls{bnn}, $\lambda = 1$}
        \label{fg:experiments:compression:fashion}
    \end{subfigure}
    \begin{subfigure}[t]{0.3\textwidth}
        \centering
        \includegraphics[height=2in]{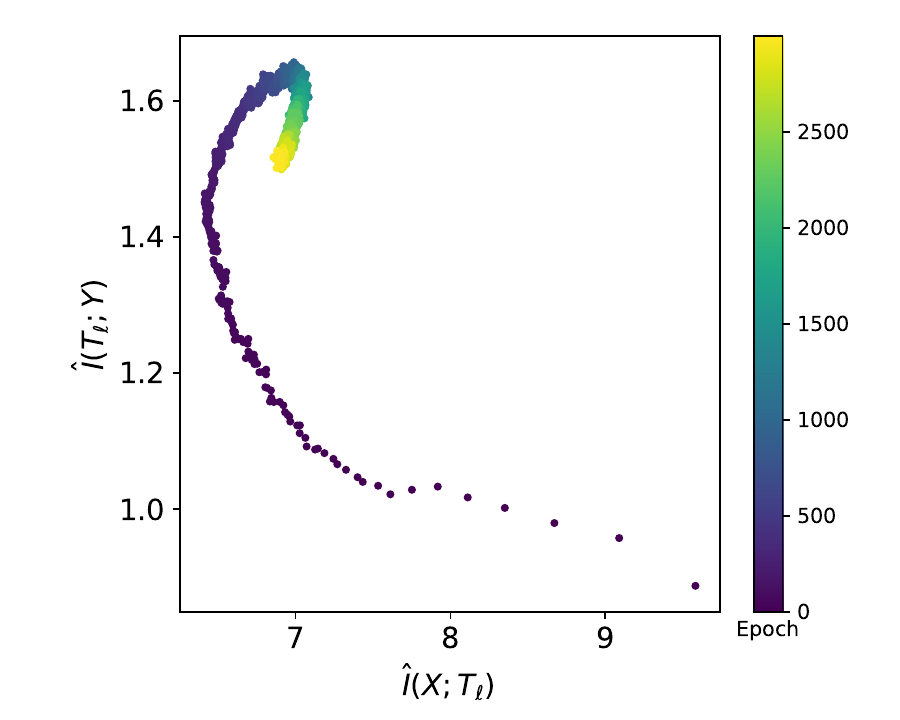}
        \caption{CIFAR-10, LeNet5, $\lambda = 0$}
        \label{fg:experiments:compression:cifar10}
    \end{subfigure}
    \caption{Select \glspl{ip} of trained models. All figures show the last (and second to last in \protect(\subref{fg:experiments:compression:fashion})) hidden \gls{fc} layer, all with $d_\ell = 10$. The achieved mean validation accuracies are \numprint[\%]{96.8}, \numprint[\%]{83.0}, and \numprint[\%]{53.3}, respectively. Compressional movement is observable in all figures, with~(\subref{fg:experiments:compression:mnist}) exhibiting prototypical compression and~(\subref{fg:experiments:compression:fashion}) making the redundancy of the last hidden layer visually observable. (\subref{fg:experiments:compression:cifar10}) shows a qualitatively different compression, as the reduction of $\hat{I}(X;T_\ell)$ happens faster than the increase of $\hat{I}(T_\ell;Y)$.}
    \label{fg:experiments:compression}
\end{figure*}

To highlight some aspects of \gls{ip} analysis, additional \glspl{ip} for select experiments are shown in \cref{fg:experiments:compression}. The Raj-like model trained on MNIST in \cref{fg:experiments:compression:mnist} exhibits strong and clear compressional movement within the first hundred epochs of training. The end-of-training value $\bar{I}(X;T_\ell)_\text{50}$ is only slightly larger than $\log_2(10)$, which indicates that the latent representations $T_\ell$ have mostly collapsed to ten binary patterns, corresponding to the ten classes of MNIST. 

The small regularised \gls{bnn} on FashionMNIST in \cref{fg:experiments:compression:fashion} still shows compression with a positive $\varrho$, but the qualitative picture is different: The last layer (left curve) only partially or temporarily compresses and, towards the end of training, increases $\hat I(X;T_3)$ towards the value $\hat I(X;T_2)$ of its preceding layer (right curve) in the \gls{ip}. This renders the third layer obsolete, as it becomes a bijective map of the previous activations\footnote{Since $T_3$ is a function of $T_2$, by the data processing inequality we can only have $H(T_3)=H(T_2)$ if the map is bijective.}, e.g., by merely permuting the preceding activation bits. While we have to look at the concrete activations to confirm this suspicion, it is interesting that this assumption can be deduced from the \gls{ip}.

\cref{fg:experiments:compression:cifar10} shows yet another qualitatively different compression phase for the LeNet5 model on CIFAR-10: Here, compression happens early during training, with the reduction of $\hat I(X;T_\ell)$ happening faster than the increase of $\hat I(T_\ell;Y)$. Considering the complexity of the underlying dataset and the model's mediocre performance, this might be caused by $\hat I(T_\ell;Y)$ only starting to increase once $\hat I(X;T_\ell)$ stops decreasing. This stands in contrast to previous work~\cite{shwartz-ziv_opening_2017,saxe_information_2019, raj_understanding_2020, lorenzen_information_2022}, among others, and might hint towards \gls{ip} analysis being able to explain mediocre performance on hard problems in general.

All selected experiments in \cref{fg:experiments:compression} have in common that the maximum value of $\hat I(X;T_\ell)$ is achieved at the first epoch. This behaviour was also visible for the \glspl{bnn} studied in~\cite{raj_understanding_2020}. In contrast, applying binning together with the plug-in estimator to deterministic neural networks with real-valued activations shows an initial increase of $\hat I(X;T_\ell)$, i.e., $\hat I(X;T_\ell)_{\max}$ is assumed at a later epoch. 
This discrepancy between \glspl{bnn} and real-valued neural networks is noteworthy and may be explained by the fact that random, small-weight initialisations result in random binary patterns in \glspl{bnn}, but in low-magnitude activations in classical neural networks, which populate only a small number of the available bins and hence lead to small entropy estimates.

\section{Generalisation of Compressed Representations}\label{sec:compression}

It was hypothesised that the late-stage compression exhibited by neural networks and, as a consequence, information-theoretically compressed latent representations, are an explanation for the excellent generalisation capabilities of deep learning. While intuitive, the scientific literature has shown mixed results regarding the connection between compression and generalisation performance~\cite[Table~1]{geiger_information_2022}. To contribute to this hypothesis, we analyse the relation between the \gls{mi} $\hat I(X;T_\ell)$ and the validation accuracy at the end of training. To this end, we aggregate $\hat{I}(X;T_\ell)$ and accuracy of the last 50 training epochs to compute Spearman's rank correlation coefficient $r_s$ over all random initialisations and weight decay coefficients for a given model architecture. 

\Cref{tb:experiments:generalisation:rank-corr} shows the resulting coefficients and $p$-values per experiment group and layer, where negative coefficients indicate that information-theoretically compressed representations lead to improved generalisation performance. As it can be seen, both positive and negative rank correlations are significant, with substantial effect sizes ($|r_s|>0.5$). This suggests that the effect of information-theoretic compression depends on the architecture and dataset, among other things.

\begin{figure*}[!t]
    \centering
    \begin{subfigure}[t]{0.3\textwidth}
        \centering
        \includegraphics[height=2in]{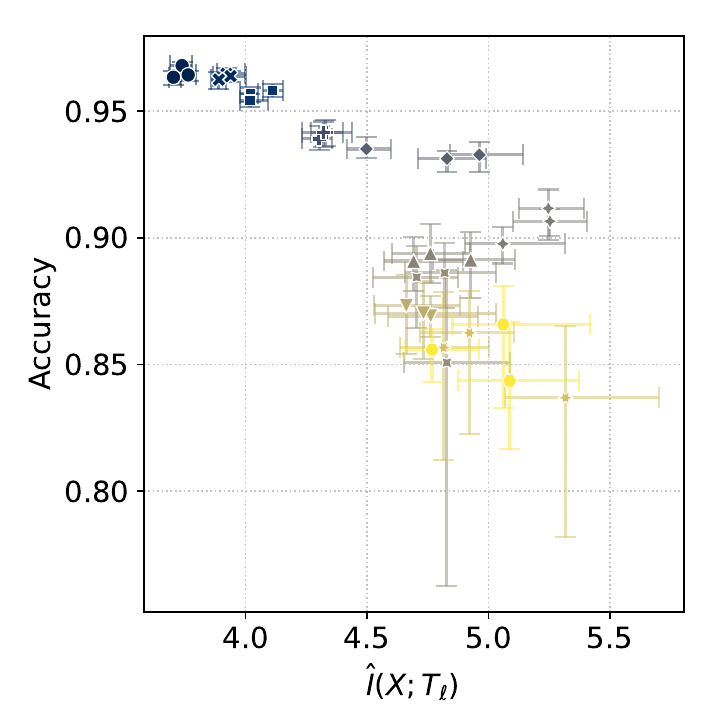}
        \caption{MNIST, Raj-like}
        \label{fg:experiments:generalisation:mnist}
    \end{subfigure}
    \begin{subfigure}[t]{0.3\textwidth}
        \centering
        \includegraphics[height=2in]{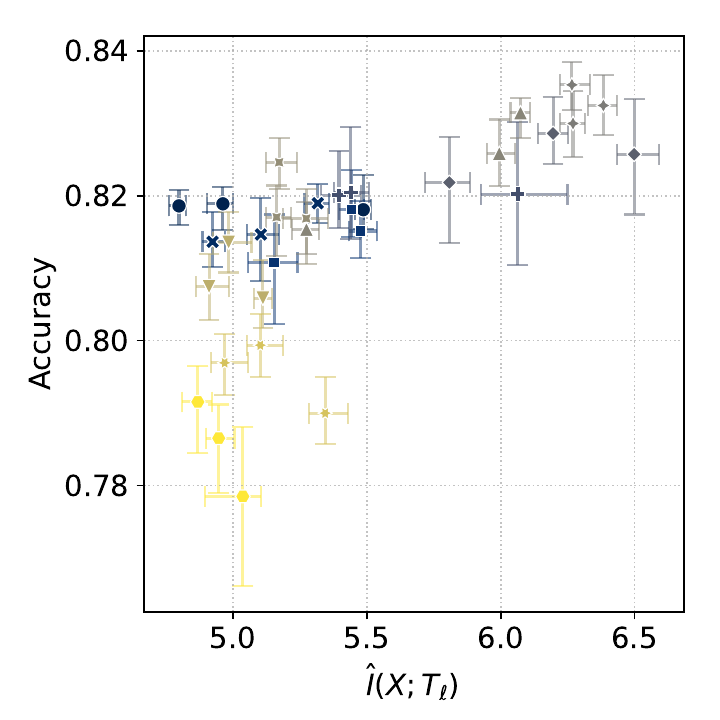}
        \caption{FashionMNIST, small \gls{bnn}}
        \label{fg:experiments:generalisation:fashion}
    \end{subfigure}
    \begin{subfigure}[t]{0.3\textwidth}
        \centering
        \includegraphics[height=2in]{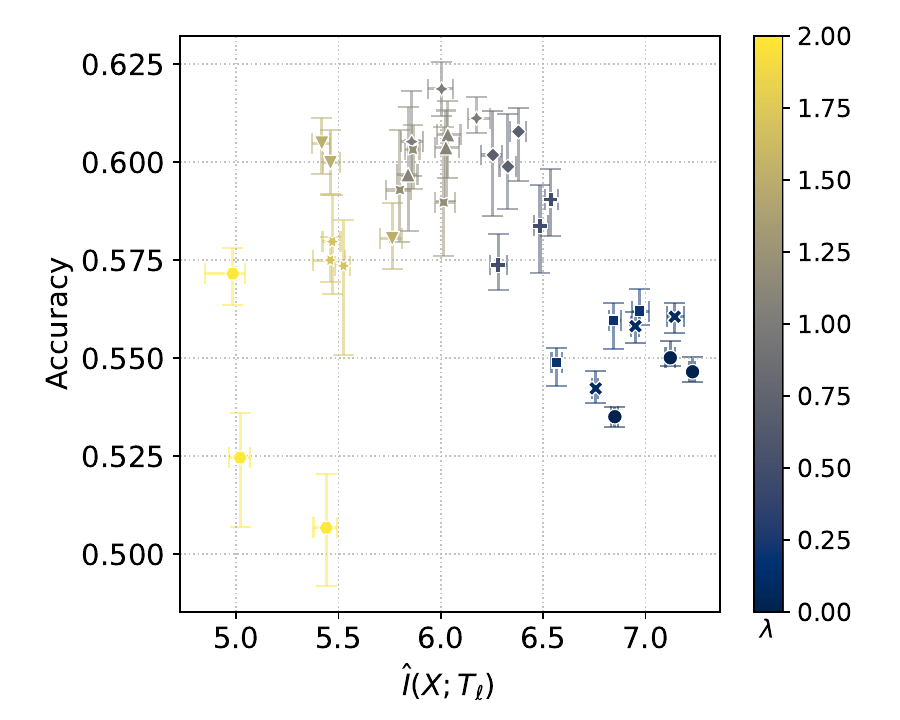}
        \caption{CIFAR-10, small LeNet5}
        \label{fg:experiments:generalisation:cifar10}
    \end{subfigure}
    \caption{Comparison of the \gls{mi} of the penultimate layer w.r.t. the input data $X$ and the validation accuracy for varying weight decay $\lambda$. Points are averaged values over the last 50 epochs, with minimum-maximum range shown within this time frame. Each figure highlights a different relation: In~(\subref{fg:experiments:generalisation:mnist}), negative correlation and a suggested transition to underfitting caused by an increased weight decay can be seen. For~(\subref{fg:experiments:generalisation:fashion}), accuracy and \gls{mi} initially increase, but start to drop after a certain threshold $\lambda$. (\subref{fg:experiments:generalisation:cifar10}) exhibits prototypical behaviour, as $\hat{I}(X;T_\ell)$ increases with $\lambda$, but accuracy initially increases and then decreases again as $\lambda$ grows larger, hinting towards a transition from a well-trained model to underfitting.}
    \label{fg:experiments:generalisation}
\end{figure*}

To shed light into these varying interplays, we plot the aggregated values over the last 50 epochs for select experiments in \cref{fg:experiments:generalisation}. The prototypical expected behaviour is seen in \cref{fg:experiments:generalisation:cifar10} for the small LeNet5 model trained on CIFAR-10: The \gls{mi} reduces with increasing weight decay parameter $\lambda$, while accuracy first increases and then decreases as $\lambda$ becomes larger. This suggests a transition from overfitting through a well-trained model to underfitting due to strong regularisation. The result is a weakly negative, but insignificant rank correlation.

On FashionMNIST, the small \gls{bnn} shows a similar behaviour regarding accuracy, see \cref{fg:experiments:generalisation:fashion}. Indeed, accuracy and \gls{mi} initially increase, but start to drop once $\lambda$ reaches a certain threshold. As a consequence, the rank correlation between these two quantities is positive and significant, indicating that compressed representations generalise worse than uncompressed latent representations.

Finally, the Raj-like model trained on MNIST, shown in \cref{fg:experiments:generalisation:mnist}, exhibits a significant negative correlation with monotonic tendency. Increasing the weight decay parameter both increases \gls{mi} and decreases accuracy, suggesting a transition into underfitting. Furthermore, the models' accuracies vary more towards end of training for large values of $\lambda$.

In summary, it appears that whether compressed representations generalise better depends at least partly on how well the regularised architecture is capable of solving the dataset-dependent task. Taking \cref{fg:experiments:generalisation:cifar10} as the prototypical behaviour, one can see that an appropriate selection of weight decay parameters can yield both positive and negative rank correlation values. The fact that in some architectures, stronger regularisation leads to larger values of $\hat I(X;T_\ell)$ is surprising and shall be the object of future study.

\begin{table}[!t]
    \nprounddigits{3}
    \centering
    \begin{threeparttable}[b]
        \caption{Spearman's rank correlation coefficients over the mean $\bar{I}(X;T_\ell)_{50}$ and the mean validation accuracy of the last 50 epochs. Positive values indicate that information-theoretic compression hurts classification performance. Each value is computed per layer over each experiment group, including weight decays. Layers are given as negative offset indices from the output layer. All layers have a width of ten neurons, except for SZT, consider \cref{tb:experiments:setup:architectures} for more detail.}
        \label{tb:experiments:generalisation:rank-corr}
        \begin{tabular}{llc|rr}
            \toprule
            Dataset & Experiment Group & Layer & $r_s$ & $p$-value \\
            \midrule
            \textit{SZT} & \textit{WD} & \textit{-4}\tnote{\S} & $\textit{\numprint{0.6019385026737969}}$ & $< \textit{0.001}$ \\
            \textit{SZT} & \textit{WD} & \textit{-3}\tnote{\S} & $\textit{\numprint{0.5477941176470589}}$ & $\textit{\numprint{0.000967498128557705}}$ \\
            SZT & WD & -2 & $\numprint{0.4953208556149733}$ & $\numprint{0.0033802458198417475}$ \\
            SZT & WD & -1 & $\numprint{0.546457219251337}$ & $\numprint{0.0010013848428244107}$ \\
            \midrule
            MNIST & Raj-like & -1 & $\numprint{-0.7613636363636364}$ & $< 0.001$ \\
            MNIST & Small BNN & -2 & $\numprint{-0.07653743315508023}$ & $\numprint{0.6720450893584659}$ \\
            MNIST & Small BNN & -1 & $\numprint{0.37800802139037437}$ & $\numprint{0.030085704875244066}$ \\
            MNIST & LeNet5 & -1 & $\numprint{-0.5725745820943481}$ & $< 0.001$ \\
            MNIST & Bottleneck & -1 & $\numprint{0.682142857142857}$ & $\numprint{0.0050863264689772515}$ \\
            MNIST & Hourglass & -1 & $\numprint{0.782142857142857}$ & $\numprint{0.0005697096277734958}$ \\
            \midrule
            FashionMNIST & Raj-like & -1 & $\numprint{-0.35995989304812837}$ & $\numprint{0.03962589120625222}$ \\
            FashionMNIST & Small BNN & -2 & $\numprint{0.5691844919786097}$ & $\numprint{0.0005466753789992462}$ \\
            FashionMNIST & Small BNN & -1 & $\numprint{0.7677139037433156}$ & $< 0.001$ \\
            FashionMNIST & LeNet5 & -1 & $\numprint{-0.28643048128342247}$ & $\numprint{0.10608596657996938}$ \\
            FashionMNIST & Bottleneck & -1 & $\numprint{0.8571428571428571}$ & $< 0.001$ \\
            FashionMNIST & Hourglass & -1 & $\numprint{0.6035714285714285}$ & $\numprint{0.01720013806879533}$ \\
            \midrule
            CIFAR-10 & LeNet5 & -1 & $\numprint{-0.6417112299465242}$ & $< 0.001$ \\
            CIFAR-10 & Variant LeNet5 & -1 & $\numprint{0.48956521739130426}$ & $\numprint{0.015176284460359705}$ \\
            CIFAR-10 & Small LeNet5 & -1 & $\numprint{-0.26169786096256686}$ & $\numprint{0.14125017625401567}$ \\
            \bottomrule
        \end{tabular}
        \begin{tablenotes}
            \item[\S] Note that the layers with eight/ten neurons are outside the suggested regime for $N \approx 800$ for reliable entropy estimation.
        \end{tablenotes}
    \end{threeparttable}
\end{table}

\section{Discussion and Conclusions}\label{sec:discussion}

In this work, we studied \gls{mi} in \glspl{bnn}, with the purpose of determining whether compression occurs during training, and whether compressed representations generalise better. While our experiments show that some form of compression during training is a prevalent phenomenon (see \cref{fg:experiments:compression}), we conclude that compressed representations are, in general, not performing better than uncompressed representations (\cref{tb:experiments:generalisation:rank-corr}). An intuitive explanation for this inconsistent picture is that a representation can be compressed not only by being cleaned of irrelevant information, but also by becoming uninformative for the task, as in underfitting models. While we believe this explanation to be valid in many settings, a closer look at \cref{fg:experiments:generalisation} shows that the interplay between compression and generalisation is more nuanced even in small \glspl{bnn}.

One obvious limitation of our study is that it is restricted to comparably small \glspl{bnn}, and then only to their smallest layers. While this is an immediate consequence of the properties of the plug-in estimator and the availability of validation data (cf.~\cref{sec:method:entropy}), we believe that slightly wider layers can be studied with more sophisticated estimators, such as the best upper bounds estimator of~\cite{paninski_estimation_2003}. Even this estimator, however, does not yield reliable estimates for analysing the \gls{ip} of contemporary networks using common datasets, suggesting that future research should focus on estimating information-theoretic quantities from high-dimensional data. A second limitation of our work is the fact that we summarise any reduction of $\hat I(X;T_\ell)$ during training via a single number $\varrho$, which is of course insufficient to capture the wide range of qualitative behaviour shown in \glspl{ip}. Specifically, while originally compression was linked to a second phase during training~\cite{shwartz-ziv_opening_2017}, our compression factor $\varrho$ is positive whenever $\hat I(X;T_\ell)$ does not achieve its maximum at the end of training, even if $\hat I(X;T_\ell)$ increases during the last epochs (e.g.,~\cref{fg:experiments:compression:fashion}). 



Our theoretical and experimental discussion of \gls{mi} estimator properties in \cref{sec:method:entropy} sheds new light on previous \gls{ip} analyses. While previous work acknowledged that deterministic networks suffer from infinite $I(X;T_\ell)$ and hence mostly display geometric properties in the \gls{ip}, our work suggests that even for \glspl{bnn}, for which $I(X;T_\ell)$ is finite, estimation may be unreliable. For example,~\cite[Fig.~3]{raj_understanding_2020} plots the \gls{ip} for a \gls{bnn} with architecture $1024-20-20-20$. Even for the later layers, the alphabet size $k=2^{20}\approx 10^{6}$ is too large to admit reliable estimation from only $N=60,000$ samples from the MNIST dataset (training and test set combined). Similarly~\cite{lorenzen_information_2022} investigated \glspl{ip} for neural networks with activations quantised to 8 bit. For a layer of width $d_\ell=8$, the corresponding alphabet size is $k=2^{d_\ell\cdot 8}\approx 10^{19}$, requiring $N\approx 10^{35}$ samples for reliable estimation. While small true values of $I(X;T_\ell)$ result in a substantially smaller effective alphabet size, i) the true value of $I(X;T_\ell)$ is neither known a priori nor can it be reasonably bounded, and ii) the fact that $\hat I(X;T_\ell)\approx \log N$ at least for early epochs and layers indicates that \gls{mi} estimation saturated (see \cref{fg:methods:ee:plug-in:suboptimality}) and that the resulting estimates are not reliable. Therefore, while restricted to only small architectures and to the specific setting of \glspl{bnn}, our work is among the first that presents reliable \gls{ip} analyses and experimental evidence regarding the connection between information-theoretic compression and generalisation.

\section*{Acknowledgments}
This research was funded by the Austrian Science Fund (FWF) under grant 10.55776/PAT7753623. Know Center is a COMET competence center that is financed by the Austrian Federal Ministry of Innovation, Mobility and Infrastructure (BMIMI), the Austrian Federal Ministry of Economy, Energy and Tourism (BMWET), the Province of Styria, the Steirische Wirtschaftsförderungsgesellschaft m.b.H. (SFG), the Vienna business agency and the Standortagentur Tirol. The COMET programme is managed by the Austrian Research Promotion Agency FFG.

\bibliographystyle{IEEEtran}
\bibliography{IEEEabrv,bibs/references_IPs_new,bibs/others}



\end{document}